\documentclass[10pt,twocolumn,letterpaper]{article}

\usepackage[pagenumbers]{cvpr}

\usepackage{graphicx}
\usepackage{amsmath,amssymb}
\usepackage{booktabs}
\usepackage{multirow}
\usepackage{xcolor}
\usepackage{colortbl}
\usepackage{microtype}

\definecolor{cvprblue}{rgb}{0.21,0.49,0.74}
\usepackage[pagebackref,breaklinks,colorlinks,allcolors=cvprblue]{hyperref}

\def\paperID{0000}
\def\confName{CVPR}
\def\confYear{2026}

\definecolor{vlmcolor}{HTML}{E65100}
\definecolor{gpucolor}{HTML}{1565C0}

\title{Out of the box age estimation through facial imagery: A Comprehensive Benchmark of Vision-Language Models vs.\ out-of-the-box Traditional Architectures}

\author{
Simiao Ren$^{1*\dagger}$, Xingyu Shen$^{1,2*}$, Ankit Raj$^{1,2}$, Albert Dai$^{2}$, Caroline (Manlin) Zhang$^{1}$, \\
Yuan Xu$^{1}$, Zexi Chen$^{3}$, Siqi Wu$^{3}$, Chen Gong$^{2}$, Yuxin Zhang$^{4}$
\vspace{2mm} \\
$^{1}$Reality Inc. \quad $^{2}$Duke University \quad $^{3}$New York University \quad $^{4}$University of Massachusetts Amherst \\
\vspace{1mm} \\
{\tt\small benren@scam.ai \{alex, ankit, albert\}@get-reality.com \quad  cg387@duke.edu} \\
{\tt\small \{mzhang1030, yuanxu.derrick\}@gmail.com \quad \{zc2610, sw5693\}@nyu.edu \quad yuxinzhang@umass.edu} \\
\vspace{1mm} \\
\small{$^*$Equal contribution \quad $^\dagger$Corresponding author}
}

\begin{document}
\maketitle
{\renewcommand{\thefootnote}{}\footnote{Authorship of this paper does not imply employment by or affiliation with Scam.AI (Reality Inc.); some authors are independent researchers who collaborated on this work.}\addtocounter{footnote}{-1}}

\begin{abstract}
Facial age estimation is critical for content moderation, age verification, and deepfake detection, yet no prior benchmark has systematically compared modern vision-language models (VLMs) against specialized age estimation architectures.
We present the first large-scale cross-paradigm benchmark, evaluating \textbf{34 models}---22 specialized architectures with publicly available pretrained weights and 12 general-purpose VLMs---across \textbf{8 standard datasets} (UTKFace, IMDB-WIKI, MORPH, AFAD, CACD, FG-NET, APPA-REAL, AgeDB) totaling 1{,}100 test images per model.
Our key finding is striking: \emph{zero-shot VLMs significantly outperform most specialized models}, achieving an average MAE of 5.65 years compared to 9.88 for non-LLM models.
The best VLM (Gemini~3 Flash Preview, MAE~4.32) outperforms the best non-LLM model (MiVOLO, MAE~5.10) by 15\%.
Only MiVOLO, which uniquely combines face and body features via Vision Transformers, competes with VLMs.
We further analyze age verification at the 18-year threshold, revealing that most non-LLM models exhibit 39--100\% false adult rates on minors while VLMs achieve 16--29\%, and demonstrate that coarse age binning (8--9 classes) consistently degrades MAE beyond 13 years.
Our stratified analysis across 14 age groups reveals that all models struggle most at extreme ages ($<$5 and 65+).
These findings challenge the assumption that task-specific architectures are necessary for age estimation and suggest that the field should redirect toward distilling VLM capabilities into efficient specialized models.
\end{abstract}

\section{Introduction}
\label{sec:intro}

Facial age estimation---predicting a person's age from a face image---is a fundamental task in computer vision with growing importance in content moderation~\cite{eu2024aiact}, age verification~\cite{ofcom2024ageverification}, and deepfake detection~\cite{ren2025llmdeepfake}.
Over the past decade, specialized architectures have progressed from handcrafted features~\cite{guo2009agewild} through convolutional neural networks~\cite{rothe2015dex,levi2015rudecarnie} to Vision Transformers~\cite{kuprashevich2024mivolo}, with each generation claiming state-of-the-art performance on standard benchmarks.

Meanwhile, general-purpose Vision-Language Models (VLMs) such as GPT-5~\cite{openai2025gpt5}, Gemini~\cite{team2025gemini}, and Claude~\cite{anthropic2025claude} have demonstrated remarkable zero-shot capabilities across diverse visual tasks, from visual question answering to optical character recognition.
These models, trained on internet-scale data with billions of image-text pairs, have acquired broad visual understanding that may encompass the age-related cues that specialized models explicitly learn.
Recent work has shown that multimodal LLMs are increasingly competitive with specialized pipelines across traditionally vision-specific tasks, including deepfake detection~\cite{ren2025llmdeepfake} and document fraud detection~\cite{liang2025llmdocument}.
A natural question arises: \emph{can these general-purpose models match or surpass specialized age estimation architectures, despite having no task-specific training?}

Prior benchmarks~\cite{paplhjak2024calltoreflect,othmani2020agesurvey} have compared traditional architectures on standard datasets, and recent work~\cite{cong2024vlmfacial} has evaluated VLMs on facial attribute \emph{classification}.
However, no existing study has systematically compared VLMs against specialized models on continuous-valued age \emph{regression} (MAE) across multiple standard datasets.
This gap is significant because (1)~the field continues to develop specialized architectures without knowing whether they match VLM baselines, and (2)~practitioners selecting models for deployment lack cross-paradigm comparisons.

We address this gap with the following contributions:
\begin{itemize}
    \item \textbf{Large-scale cross-paradigm benchmark}: We evaluate 34 models (22 specialized architectures with publicly available pretrained weights and 12 zero-shot VLMs) on 8 standard age estimation datasets, the largest such comparison to date.
    \item \textbf{VLM dominance}: We show that zero-shot VLMs occupy 12 of the top 13 positions, with an average MAE of 5.65 vs.\ 9.88 for non-LLM models---a 43\% improvement without any task-specific training.
    \item \textbf{Age verification analysis}: We evaluate all models at the critical 18-year threshold, revealing that most non-LLM models have unacceptably high false adult rates (39--100\%) while VLMs achieve substantially lower rates (16--29\%).
    \item \textbf{Coarse binning failure mode}: We demonstrate that models using 8--9 discrete age bins consistently produce MAE $>$13 years, quantifying the penalty of discretization.
    \item \textbf{Age-group stratified analysis}: We provide per-age-group MAE breakdowns revealing that all models struggle at extreme ages, with VLMs degrading more gracefully.
\end{itemize}

The remainder of this paper is organized as follows: \Cref{sec:related} reviews related work, \Cref{sec:benchmark} describes our benchmark design, \Cref{sec:results} presents results and analysis, \Cref{sec:discussion} discusses implications, and \Cref{sec:conclusion} concludes.

\section{Related Work}
\label{sec:related}

\paragraph{Traditional Age Estimation.}
Early approaches relied on handcrafted features such as Active Appearance Models and Bio-Inspired Features~\cite{guo2009agewild}.
The deep learning era began with CNN-based regression~\cite{levi2015rudecarnie} and the Deep EXpectation (DEX) method~\cite{rothe2015dex}, which recast age estimation as classification over 101 age classes followed by expected value computation, achieving strong results on the IMDB-WIKI dataset~\cite{rothe2018imdbwiki}.
Subsequent work explored ordinal regression~\cite{niu2016afad,cao2020coral}, mean-variance loss~\cite{pan2018meanvariance}, lightweight architectures such as SSR-Net~\cite{yang2018ssrnet}, and multi-task frameworks combining age with gender and identity~\cite{serengil2024deepface,deng2022insightface}.
More recently, Vision Transformers~\cite{dosovitskiy2021vit} and foundation models~\cite{ren2024segmentspace} have reshaped visual understanding tasks, with MiVOLO~\cite{kuprashevich2024mivolo} achieving state-of-the-art age estimation results by jointly processing face and body crops via a ViT backbone with YOLOv8~\cite{jocher2023yolov8} detection.
Othmani~\etal~\cite{othmani2020agesurvey} showed that face-related pretraining~\cite{parkhi2015vggface,cao2018vggface2} and architectures like Xception~\cite{chollet2017xception} can achieve MAE as low as 2.01 on MORPH with CASIA-Web pretraining, highlighting the importance of domain-specific transfer learning.
Bekhouche~\etal~\cite{bekhouche2024msdnn} recently proposed a multi-stage deep neural network achieving MAE of 2.59 on MORPH-II and 4.90 on CACD when trained on full datasets with optimized splits---demonstrating that specialized architectures continue to advance, though such results are not directly comparable to our sampled evaluation of pretrained models.
Beyond classification and regression, retrieval-based methods have also been explored: Shin~\etal~\cite{shin2022mwr} proposed Moving Window Regression (MWR), which iteratively refines age predictions via $k$-NN retrieval over a reference gallery, achieving competitive results on in-distribution data but with predictions bounded by the reference set's age range.

\paragraph{Existing Benchmarks.}
Angulu~\etal~\cite{angulu2018agesurvey} surveyed age estimation methods up to 2018, while Othmani~\etal~\cite{othmani2020agesurvey} provided a more comprehensive comparative analysis.
Most recently, Paplham and Franc~\cite{paplhjak2024calltoreflect} presented a ``Call to Reflect'' at CVPR 2024, noting inconsistencies in evaluation protocols across the literature and proposing a unified benchmark.
However, their benchmark exclusively evaluates traditional architectures and does not include VLMs, leaving a significant gap in understanding how general-purpose models compare to specialized ones.

\paragraph{VLMs for Facial Analysis.}
Cong~\etal~\cite{cong2024vlmfacial} evaluated VLMs on facial attribute understanding, but focused on \emph{classification} tasks (e.g., ``young vs.\ old'') rather than continuous age regression.
Their finding that VLMs can recognize facial attributes motivated our investigation into whether this capability extends to fine-grained age estimation.

Several concurrent works have explored VLMs for facial age estimation.
Sun~\etal~\cite{sun2024facemllm} propose Face-MLLM, a specialized face perception model achieving MAE~5.06 on AgeDB.
Hassanpour~\etal~\cite{hassanpour2024chatgpt} evaluate GPT-4 on age estimation and find ``reasonable accuracy.''
Malik~\etal~\cite{malik2025gras} audit demographic bias in VLM age recognition across gender, race, and skin tone.
Jamo~\cite{jamo2025resolution} examines the impact of image resolution on DeepFace and InsightFace age estimation.
Our work differs from these studies in scale (34 models across 8 datasets) and in providing a direct cross-paradigm comparison against a comprehensive set of specialized architectures rather than evaluating VLMs in isolation.
To our knowledge, ours is the first \emph{large-scale cross-paradigm} benchmark for continuous age estimation.

\paragraph{Age Verification and Regulatory Context.}
The EU AI Act~\cite{eu2024aiact} and UK Ofcom guidelines~\cite{ofcom2024ageverification} increasingly mandate age verification for online platforms, creating practical demand for accurate age estimation at decision thresholds (typically 13, 16, and 18 years).
This motivates our threshold analysis in \Cref{sec:threshold}, which evaluates models not just on MAE but on their ability to correctly classify individuals as above or below the legal age boundary.

\section{Benchmark Design}
\label{sec:benchmark}

\subsection{Models}
\label{sec:models}

We evaluate 34 models spanning two paradigms: 22 specialized (non-LLM) architectures and 12 general-purpose VLMs.
\Cref{tab:taxonomy} summarizes the model taxonomy.

\textbf{Inclusion criterion.} For the non-LLM models, we restrict our benchmark to architectures that provide \emph{publicly available pretrained weights}---i.e., models that can be downloaded and run for inference without retraining.
We exclude models whose original weights are no longer available (e.g., dead Google Drive links), models that only provide training code without pretrained checkpoints, and models whose weights were trained on fewer than 5{,}000 images.
This criterion ensures that our evaluation reflects each architecture's published capability rather than an ad hoc retraining on potentially mismatched data.
Several well-known methods with publicly available code were excluded under this criterion: FP-Age~\cite{huang2021fpage} (weight server offline), CORAL-CNN~\cite{cao2020coral} (Google Drive weights deleted), and the Inception-based model of Levi and Hassner~\cite{levi2015rudecarnie} (hosted weights no longer accessible).
These architectures are extensively benchmarked under controlled cross-dataset settings by Paplham and Franc~\cite{paplhjak2024calltoreflect}, who demonstrate that evaluation methodology---particularly facial alignment and training data---often matters more than architecture choice; we refer readers to their work for a comprehensive comparison of these methods.

\textbf{Non-LLM models} cover the major architectural families in age estimation:
(1)~\emph{CNN-Regression}: direct age regression from CNN features (e.g., ResNet~\cite{he2016resnet}, SE-ResNet~\cite{hu2018senet}), including models with specialized losses such as mean-variance loss~\cite{pan2018meanvariance};
(2)~\emph{CNN-Expected Value}: classification over fine-grained age bins followed by expected value computation, exemplified by DEX~\cite{rothe2015dex} with VGG-16~\cite{simonyan2015vgg};
(3)~\emph{Vision Transformer}: MiVOLO~\cite{kuprashevich2024mivolo}, which combines ViT with YOLOv8 face/body detection;
(4)~\emph{Multi-task}: models jointly predicting age, gender, and/or identity (DeepFace~\cite{serengil2024deepface}, InsightFace~\cite{deng2022insightface});
(5)~\emph{Lightweight}: compact architectures for deployment (SSR-Net~\cite{yang2018ssrnet});
(6)~\emph{Retrieval-based}: MWR~\cite{shin2022mwr}, which uses $k$-NN retrieval over a reference gallery followed by iterative window regression;
(7)~\emph{Coarse-bin}: models using 8--9 discrete age categories~\cite{facexformer2024}.

\textbf{VLMs} are evaluated in a zero-shot setting via the OpenRouter API (accessed February~6--7, 2026) using a standardized prompt: ``\emph{Estimate the age of the person in this photograph. Respond with ONLY a single integer representing their age in years. Do not include any other text, explanation, or units.}''
Temperature is set to 0 for deterministic outputs; maximum token limit is 256.
We evaluate models from 9 providers: Google (Gemini), OpenAI (GPT-5), Anthropic (Claude), Alibaba (Qwen), Meta (Llama), Mistral, xAI (Grok), ByteDance (Seed), and Moonshot (Kimi).
Model version strings are recorded in our code repository for reproducibility.

\begin{table}[t]
\centering
\caption{Model taxonomy. We evaluate 22 specialized (non-LLM) architectures---restricted to those with publicly available pretrained weights---across 7 categories and 12 general-purpose VLMs in a zero-shot setting.}
\label{tab:taxonomy}
\small
\setlength{\tabcolsep}{3pt}
\begin{tabular}{@{}llc@{}}
\toprule
\textbf{Category} & \textbf{Representative Models} & \textbf{\#} \\
\midrule
\multicolumn{3}{@{}l}{\textit{Specialized (Non-LLM) Models}} \\
\midrule
CNN-Regression & Yu4u, BoyuanJiang, Py-Agender & 10 \\
CNN-Expected Value & DEX (VGG-16) & 1 \\
Vision Transformer & MiVOLO (ViT + YOLOv8) & 1 \\
Multi-task & DeepFace, InsightFace, Muno-AI & 4 \\
Lightweight & SSR-Net & 1 \\
Retrieval-based & MWR (VGG-16 + $k$-NN) & 1 \\
Coarse-bin (8--9 cls) & FaceXFormer, ChienThan & 4 \\
\midrule
\multicolumn{3}{@{}l}{\textit{Zero-shot Vision-Language Models}} \\
\midrule
VLM & Gemini, GPT-5, Claude, Qwen, & 12 \\
    & Llama, Mistral, Grok, Seed, Kimi & \\
\midrule
\textbf{Total} & & \textbf{34} \\
\bottomrule
\end{tabular}
\end{table}

\subsection{Datasets}
\label{sec:datasets}

We evaluate on 8 widely-used age estimation datasets spanning diverse demographics, imaging conditions, and age ranges (\Cref{tab:datasets}).
For each dataset, we randomly sample 100 images (seed~42), except AgeDB~\cite{moschoglou2017agedb} where we sample 400 images, to ensure tractable evaluation across all 34 models while maintaining statistical reliability (1{,}100 images per model).

\begin{table}[t]
\centering
\caption{Datasets used in our benchmark. We randomly sample 100 images per dataset (seed 42), except AgeDB which uses 400 images.}
\label{tab:datasets}
\small
\setlength{\tabcolsep}{3pt}
\begin{tabular}{@{}lcrl@{}}
\toprule
\textbf{Dataset} & \textbf{Size} & \textbf{Ages} & \textbf{Description} \\
\midrule
UTKFace~\cite{zhang2017utkface} & 24K & 0--116 & In-the-wild faces \\
IMDB-WIKI~\cite{rothe2018imdbwiki} & 213K & 0--100 & Celebrity images \\
MORPH~\cite{ricanek2006morph} & 50K & 16--77 & Mugshot-style \\
AFAD~\cite{niu2016afad} & 165K & 15--72 & Asian Face Age \\
CACD~\cite{chen2015cacd} & 155K & 14--62 & Cross-Age Celebrity \\
FG-NET~\cite{panis2016fgnet} & 1K & 0--69 & Fine-grained aging \\
APPA-REAL~\cite{agustsson2017apparent} & 7.5K & 0--95 & Apparent age labels \\
AgeDB~\cite{moschoglou2017agedb} & 16K & 1--101 & In-the-wild, labeled \\
\bottomrule
\end{tabular}
\end{table}

\subsection{Evaluation Protocol}
\label{sec:evaluation}

\textbf{Primary metric.} Mean Absolute Error (MAE) in years: $\text{MAE} = \frac{1}{N}\sum_{i=1}^{N}|y_i - \hat{y}_i|$, where $y_i$ is the ground-truth age and $\hat{y}_i$ is the predicted age.

\textbf{Threshold metrics.} For age verification analysis (\Cref{sec:threshold}), we compute:
(1)~\emph{False Adult Rate} (FAR): percentage of actual minors ($y < 18$) predicted as adults ($\hat{y} \geq 18$);
(2)~\emph{False Minor Rate} (FMR): percentage of actual adults ($y \geq 18$) predicted as minors ($\hat{y} < 18$).

\textbf{Stratified analysis.} We compute per-age-group MAE using 5-year bins (0--4, 5--9, \ldots, 65+) to identify age ranges where models systematically fail.

\textbf{Unified format.} All predictions are stored in standardized CSV files with columns \texttt{image\_path}, \texttt{label\_age}, and \texttt{predicted\_age}, enabling reproducible analysis.

\section{Results}
\label{sec:results}

\subsection{Overall Performance}
\label{sec:overall}

\Cref{tab:main} presents the full results across all 8 datasets (showing the top 17 and bottom 5).
The finding is unambiguous: \textbf{zero-shot VLMs dominate}.
VLMs occupy 12 of the top 13 positions, achieving an average MAE of 5.65 years compared to 9.88 for non-LLM models---a 43\% reduction in error.

To ensure a fair comparison, we also compute the non-LLM average \emph{excluding} the 4 coarse-bin models (which suffer from a fundamental discretization floor; see \Cref{sec:binning}).
This yields a non-LLM average of 8.74---still 55\% higher than the VLM average.
Restricting further to only the top~10 non-LLM models (those with MAE~$<$~9.0), the average is 7.48, still 32\% higher than VLMs (\Cref{tab:architecture}).

\begin{figure}[t]
\centering
\includegraphics[width=\linewidth]{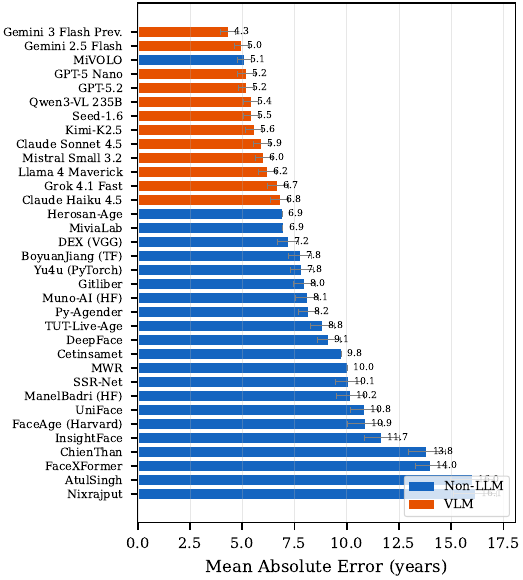}
\caption{All 34 models ranked by average MAE. \textcolor{vlmcolor}{Orange} = VLM, \textcolor{gpucolor}{blue} = non-LLM. VLMs occupy 12 of the top 13 positions.}
\label{fig:ranked}
\end{figure}

\begin{figure}[t]
\centering
\includegraphics[width=\linewidth]{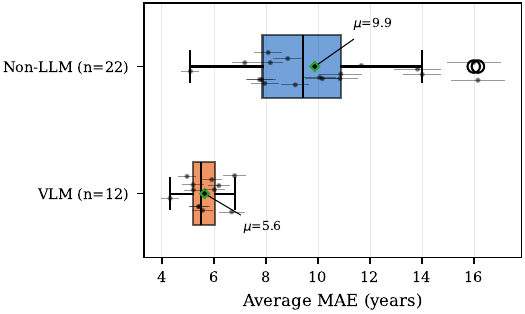}
\caption{Distribution of average MAE by model type. The VLM interquartile range falls entirely below the non-LLM median.}
\label{fig:boxplot}
\end{figure}

\Cref{fig:ranked} shows all 34 models ranked by average MAE.
The performance gap between VLMs and non-LLM models is visually striking, with a clear separation around rank 13.
\Cref{fig:boxplot} further illustrates this distribution gap: the entire VLM interquartile range (5.21--6.06) falls below the non-LLM median (9.45).

The best overall model is \textbf{Gemini~3 Flash Preview} (MAE~4.32), followed by Gemini~2.5 Flash (4.97) and \textbf{MiVOLO} (5.10).
The best non-LLM model, MiVOLO (MAE~5.10), ranks 3rd overall---the only specialized model competitive with top VLMs.
A Mann-Whitney~$U$ test confirms the VLM--non-LLM gap is statistically significant ($U = 9.0$, $p < 10^{-5}$).
Bootstrap 95\% confidence intervals (10{,}000 resamples) on per-model average MAE range from $\pm$0.34 for the best VLMs to $\pm$1.03 for high-variance coarse-bin models, indicating that the ranking is robust to sampling variation.

\begin{figure*}[t]
\centering
\includegraphics[width=\textwidth]{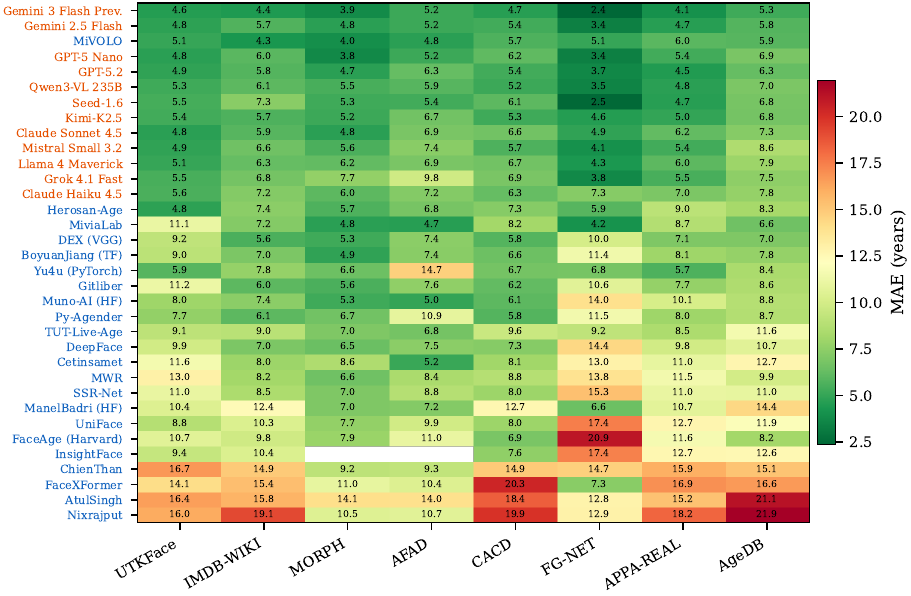}
\caption{Complete MAE heatmap across all 34 models and 8 datasets. Models sorted by average MAE (top = best). \textcolor{vlmcolor}{Orange} labels = VLM, \textcolor{gpucolor}{blue} labels = non-LLM. Darker cells indicate higher error. The top tier is dominated by VLMs with uniformly low errors across all datasets.}
\label{fig:heatmap}
\end{figure*}

\Cref{fig:heatmap} provides a detailed view of per-dataset MAE for all 34 models.
Notable dataset-specific patterns include:
(1)~\textbf{FG-NET} exhibits the widest variance (2.36--20.94 MAE), as it contains very young children that many non-LLM models trained on adult-centric datasets fail to handle;
(2)~\textbf{MORPH} favors non-LLM models more than other datasets: MiVOLO achieves 4.02, likely due to MORPH's controlled imaging conditions matching its training distribution;
(3)~\textbf{CACD} and \textbf{IMDB-WIKI} show the largest VLM advantage, suggesting that celebrity images with rich contextual cues (red carpet events, professional photos) particularly benefit VLMs' holistic understanding;
(4)~\textbf{AgeDB} produces generally higher errors across all models (best: 5.32 MAE by Gemini~3 Flash), reflecting its challenging in-the-wild conditions and broad age range (1--101).

\begin{table*}[t]
\centering
\caption{Main results: MAE (years, $\downarrow$) across 8 datasets. Models sorted by average MAE. \textcolor{vlmcolor}{Orange} = VLM, \textcolor{gpucolor}{Blue} = non-LLM. \textbf{Bold} = best per column; \underline{underline} = second best. Only non-LLM models with publicly available pretrained weights are included.}
\label{tab:main}
\small
\setlength{\tabcolsep}{3.5pt}
\begin{tabular}{@{}rlccccccccc|c@{}}
\toprule
\textbf{\#} & \textbf{Model} & \textbf{UTK} & \textbf{IMDB} & \textbf{MORPH} & \textbf{AFAD} & \textbf{CACD} & \textbf{FG-NET} & \textbf{APPA} & \textbf{AgeDB} & \textbf{Avg.} \\
\midrule
1 & \textcolor{vlmcolor}{Gemini 3 Flash Prev.} & \textbf{4.62} & \underline{4.43} & \underline{3.85} & 5.23 & \textbf{4.67} & \textbf{2.36} & \textbf{4.09} & \textbf{5.32} & \textbf{4.32} \\
2 & \textcolor{vlmcolor}{Gemini 2.5 Flash} & 4.81 & 5.74 & 4.75 & 5.22 & 5.38 & 3.42 & 4.67 & \underline{5.80} & \underline{4.97} \\
3 & \textcolor{gpucolor}{MiVOLO} & 5.07 & \textbf{4.32} & 4.02 & \underline{4.77} & 5.65 & 5.09 & 6.00 & 5.88 & 5.10 \\
4 & \textcolor{vlmcolor}{GPT-5 Nano} & 4.80 & 5.96 & \textbf{3.75} & 5.22 & 6.15 & 3.41 & 5.40 & 6.90 & 5.20 \\
5 & \textcolor{vlmcolor}{GPT-5.2} & 4.94 & 5.81 & 4.74 & 6.29 & 5.41 & 3.66 & \underline{4.54} & 6.26 & 5.21 \\
6 & \textcolor{vlmcolor}{Qwen3-VL 235B} & 5.32 & 6.09 & 5.50 & 5.90 & \underline{5.19} & 3.53 & 4.80 & 6.97 & 5.41 \\
7 & \textcolor{vlmcolor}{Seed-1.6} & 5.47 & 7.33 & 5.32 & 5.39 & 6.09 & \underline{2.50} & 4.66 & 6.82 & 5.45 \\
8 & \textcolor{vlmcolor}{Kimi-K2.5} & 5.39 & 5.69 & 5.20 & 6.65 & 5.28 & 4.61 & 4.96 & 6.76 & 5.57 \\
9 & \textcolor{vlmcolor}{Claude Sonnet 4.5} & \underline{4.76} & 5.89 & 4.78 & 6.87 & 6.64 & 4.92 & 6.19 & 7.29 & 5.92 \\
10 & \textcolor{vlmcolor}{Mistral Small 3.2} & 4.89 & 6.58 & 5.60 & 7.36 & 5.69 & 4.06 & 5.40 & 8.59 & 6.02 \\
11 & \textcolor{vlmcolor}{Llama 4 Maverick} & 5.08 & 6.31 & 6.24 & 6.93 & 6.65 & 4.35 & 6.05 & 7.94 & 6.19 \\
12 & \textcolor{vlmcolor}{Grok 4.1 Fast} & 5.54 & 6.77 & 7.69 & 9.81 & 6.91 & 3.79 & 5.50 & 7.55 & 6.69 \\
13 & \textcolor{vlmcolor}{Claude Haiku 4.5} & 5.64 & 7.19 & 6.05 & 7.18 & 6.32 & 7.26 & 6.99 & 7.75 & 6.80 \\
\midrule
14 & \textcolor{gpucolor}{Herosan-Age} & 4.83 & 7.41 & 5.69 & 6.80 & 7.26 & 5.92 & 9.04 & 8.29 & 6.91 \\
15 & \textcolor{gpucolor}{Mivialab-Age} & 11.12 & 7.21 & 4.78 & \textbf{4.69} & 8.18 & 4.15 & 8.73 & 6.63 & 6.94 \\
16 & \textcolor{gpucolor}{DEX (VGG)} & 9.22 & 5.64 & 5.31 & 7.41 & 5.84 & 10.01 & 7.07 & 7.04 & 7.19 \\
17 & \textcolor{gpucolor}{BoyuanJiang (TF)} & 9.03 & 7.00 & 4.91 & 7.38 & 6.56 & 11.38 & 8.05 & 7.76 & 7.76 \\
\midrule
\multicolumn{2}{@{}l}{\textit{Bottom 5 (all non-LLM):}} & & & & & & & & & \\
30 & \textcolor{gpucolor}{InsightFace} & 9.42 & 10.36 & --- & --- & 7.61 & 17.39 & 12.66 & 12.57 & 11.67 \\
31 & \textcolor{gpucolor}{ChienThan} & 16.67 & 14.92 & 9.15 & 9.29 & 14.94 & 14.70 & 15.88 & 15.12 & 13.83 \\
32 & \textcolor{gpucolor}{FaceXFormer} & 14.13 & 15.39 & 10.98 & 10.42 & 20.31 & 7.35 & 16.88 & 16.59 & 14.01 \\
33 & \textcolor{gpucolor}{AtulSingh} & 16.45 & 15.79 & 14.11 & 14.00 & 18.40 & 12.85 & 15.22 & 21.07 & 15.99 \\
34 & \textcolor{gpucolor}{Nixrajput} & 16.04 & 19.14 & 10.46 & 10.66 & 19.86 & 12.91 & 18.18 & 21.91 & 16.14 \\
\bottomrule
\end{tabular}
\end{table*}

\begin{table}[t]
\centering
\caption{Architecture category analysis. Mean and standard deviation of average MAE within each category. Coarse-bin models (8--9 classes) consistently produce the highest errors. ``Non-LLM excl.\ coarse'' excludes the 4 coarse-bin models.}
\label{tab:architecture}
\small
\setlength{\tabcolsep}{4pt}
\begin{tabular}{@{}lccc@{}}
\toprule
\textbf{Category} & \textbf{\#} & \textbf{Mean MAE} & \textbf{Std.} \\
\midrule
VLM (zero-shot) & 12 & 5.65 & 0.72 \\
ViT (MiVOLO) & 1 & 5.10 & --- \\
CNN-Regression & 10 & 8.52 & 1.36 \\
CNN-Expected Value & 1 & 7.19 & --- \\
Multi-task & 4 & 9.94 & 1.62 \\
Retrieval (MWR) & 1 & 10.04 & --- \\
Lightweight (SSR) & 1 & 10.07 & --- \\
Coarse-bin (8--9 cls) & 4 & \textbf{15.00} & 1.25 \\
\midrule
Non-LLM excl.\ coarse & 18 & 8.74 & 1.71 \\
\bottomrule
\end{tabular}
\end{table}

\subsection{Why MiVOLO Stands Alone}
\label{sec:mivolo}

MiVOLO~\cite{kuprashevich2024mivolo} is the only non-LLM model in the top 5.
Its success stems from three design choices absent in other non-LLM models:
(1)~a \textbf{Vision Transformer} backbone rather than CNN, enabling global attention over face features;
(2)~\textbf{joint face and body} processing via YOLOv8~\cite{jocher2023yolov8} detection, providing contextual cues (clothing, posture) that supplement facial appearance;
(3)~training on a \textbf{large-scale proprietary dataset} with diverse demographics.

The second-best non-LLM model, Herosan-Age (MAE~6.91), uses ResNet-18 with mean-variance loss~\cite{pan2018meanvariance} but lacks body context or transformer attention.
The gap between MiVOLO (5.10) and Herosan-Age (6.91)---1.81 years---is larger than the gap between MiVOLO and the best VLM (0.78 years), underscoring MiVOLO's outlier status among specialized models.

\subsection{The Coarse Binning Problem}
\label{sec:binning}

Four models in our benchmark use 8--9 discrete age categories (e.g., ``0--2'', ``4--6'', ``8--12'', ``15--20'', ``25--32'', ``38--43'', ``48--53'', ``60+''): FaceXFormer, ChienThan, AtulSingh, and Nixrajput.
\Cref{tab:architecture} shows that these models collectively achieve a mean MAE of 15.00, compared to 8.52 for CNN-regression models---a 76\% increase in error attributable primarily to discretization.

\begin{figure}[t]
\centering
\includegraphics[width=\linewidth]{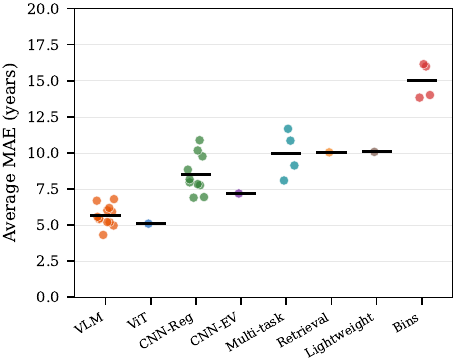}
\caption{MAE by architecture category. Each dot is one model; black bars show category means. Coarse-bin models are uniformly poor. VLMs show the tightest cluster with lowest mean.}
\label{fig:arch_scatter}
\end{figure}

\Cref{fig:arch_scatter} visualizes the MAE distribution across architecture categories.
The coarse-bin cluster is tightly concentrated at the bottom of the ranking, with no model achieving MAE below 13.6.
The fundamental issue is mathematical: with 8 bins spanning 0--100 years, the expected discretization error alone is approximately 6 years, establishing a hard floor that no amount of improved classification accuracy can overcome.

\subsection{Age Verification at the 18-Year Threshold}
\label{sec:threshold}

For content moderation and age verification, the critical question is not just MAE but whether a model correctly classifies individuals as above or below a legal age boundary.
\Cref{tab:threshold} presents False Adult Rate (FAR, minors predicted as adults) and False Minor Rate (FMR, adults predicted as minors) at the 18-year threshold across all datasets.

The results reveal a stark safety gap.
Most non-LLM models exhibit FAR rates of 39--100\%, meaning they would \emph{fail to identify the majority of minors} when used for age verification.
In contrast, VLMs achieve FAR rates of 16--29\%, representing a 2--5$\times$ improvement.
MiVOLO again stands as the best non-LLM model (FAR~21.2\%), comparable to mid-tier VLMs.

\begin{figure}[t]
\centering
\includegraphics[width=\linewidth]{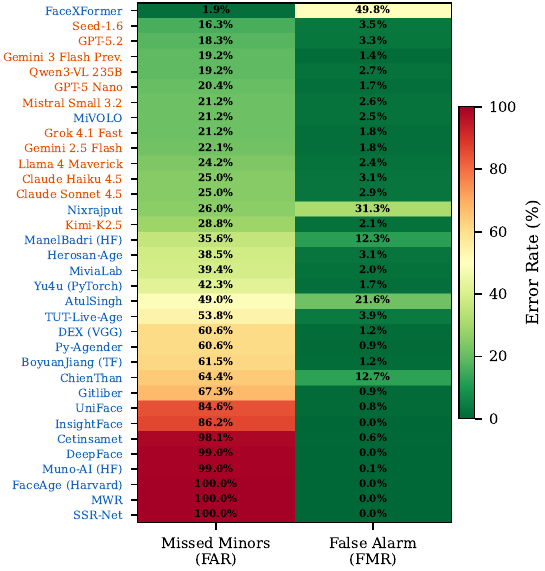}
\caption{Error rates at the 18-year threshold. Left: False Adult Rate (minors missed); Right: False Minor Rate. Models sorted by FAR. VLM model names in \textcolor{vlmcolor}{orange}.}
\label{fig:threshold}
\end{figure}

\Cref{fig:threshold} visualizes these error rates as a heatmap sorted by FAR.
A notable outlier is FaceXFormer, which achieves the lowest FAR (1.9\%) but at the cost of an extreme FMR (49.8\%)---it effectively predicts nearly all individuals as minors, rendering it useless for practical age verification.
Excluding this degenerate case, the pattern is clear: VLMs cluster near the top with consistently low error rates across both columns, while non-LLM models dominate the bottom rows with high FAR values---a pattern directly attributable to their higher MAE around the threshold age range.

\begin{table}[t]
\centering
\caption{Age verification at the 18-year threshold across all datasets. FAR = False Adult Rate (\% of actual minors predicted $\geq$18); FMR = False Minor Rate (\% of actual adults predicted $<$18). Lower FAR/FMR is better. Selected models with $\geq$5 minor samples shown.}
\label{tab:threshold}
\small
\setlength{\tabcolsep}{4pt}
\begin{tabular}{@{}lrrc@{}}
\toprule
\textbf{Model} & \textbf{FAR}$\downarrow$ & \textbf{FMR}$\downarrow$ & \textbf{Type} \\
\midrule
\textcolor{vlmcolor}{Seed-1.6} & 16.3 & 3.5 & VLM \\
\textcolor{vlmcolor}{Gemini 3 Flash Prev.} & 19.2 & 1.4 & VLM \\
\textcolor{vlmcolor}{Qwen3-VL 235B} & 19.2 & 2.7 & VLM \\
\textcolor{vlmcolor}{GPT-5 Nano} & 20.4 & 1.7 & VLM \\
\textcolor{vlmcolor}{Mistral Small 3.2} & 21.2 & 2.6 & VLM \\
\textcolor{gpucolor}{MiVOLO} & 21.2 & 2.5 & Non-LLM \\
\textcolor{vlmcolor}{Gemini 2.5 Flash} & 22.1 & 1.8 & VLM \\
\textcolor{vlmcolor}{Claude Sonnet 4.5} & 25.0 & 2.9 & VLM \\
\textcolor{vlmcolor}{Kimi-K2.5} & 28.8 & 2.1 & VLM \\
\midrule
\textcolor{gpucolor}{Herosan-Age} & 38.5 & 3.1 & Non-LLM \\
\textcolor{gpucolor}{DEX (VGG)} & 60.6 & 1.2 & Non-LLM \\
\textcolor{gpucolor}{InsightFace} & 86.2 & 0.0 & Non-LLM \\
\textcolor{gpucolor}{DeepFace} & 99.0 & 0.0 & Non-LLM \\
\textcolor{gpucolor}{SSR-Net} & 100.0 & 0.0 & Non-LLM \\
\bottomrule
\end{tabular}
\end{table}



\subsection{Age-Group Stratified Analysis}
\label{sec:stratified}

\begin{figure}[t]
\centering
\includegraphics[width=\linewidth]{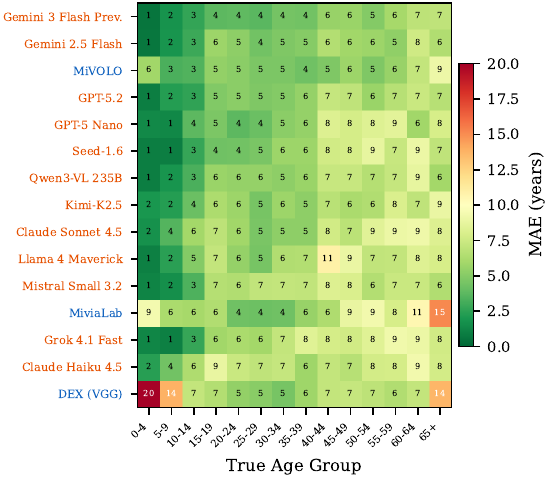}
\caption{Per-age-group MAE heatmap for top 15 models. Darker = higher error. All models struggle at extreme ages, but VLMs (\textcolor{vlmcolor}{orange} labels) degrade more gracefully.}
\label{fig:age_bin_mae}
\end{figure}

\Cref{fig:age_bin_mae} shows per-age-group MAE for the top 15 models, revealing distinct performance patterns across the human lifespan. 

(1) \textbf{Extreme Age Groups ($<$5 and 65+):} These ranges are universally difficult, with even top-tier VLMs producing MAE of 5--10 years. This performance degradation likely stems from a \textbf{long-tail data distribution} issue: in typical internet-scale pretraining sets and social media, images of infants and the elderly are significantly underrepresented compared to young adults.

(2) \textbf{Young Adults (20--34):} This cohort is the ``sweet spot'' for all models, with top VLMs achieving MAE $<$3 years. We hypothesize this is due to massive exposure in training data, as this age group is the most active on digital platforms and social media. Furthermore, individuals in this range rarely undergo significant age-altering cosmetic procedures, preserving natural aging cues.

(3) \textbf{Middle-aged Adults (35--64) and the VLM Divergence:} An interesting divergence emerges here: VLMs maintain a stable MAE of 4--6 years, while non-LLM models often exceed 10 years. We propose two complementary hypotheses for this VLM superiority:
\begin{itemize}
    \item \textbf{Contextual Reasoning:} VLMs can leverage non-facial cues (e.g., gray hair, subtle wrinkles, professional attire, or ``red carpet'' backgrounds in celebrity datasets like IMDB-WIKI) to infer age holistically. 
    \item \textbf{Robustness to Cosmetic Alterations:} This age group is more likely to undergo cosmetic surgery or use heavy anti-aging treatments. While such alterations may confuse non-LLM models that rely strictly on low-level facial textures, VLMs may ``cheat'' by associating the person's identity with known biographical timelines or by identifying the specific visual style of different eras, effectively bypassing the deceptive effects of cosmetic procedures.
\end{itemize}

\section{Discussion}
\label{sec:discussion}

\paragraph{Why VLMs Succeed.}
We hypothesize three factors driving VLM superiority:
(1)~\textbf{Scale of pretraining}: VLMs are trained on internet-scale data encompassing millions of captioned face images with implicit age information (``grandmother'', ``toddler'', ``college student''), providing a richer training signal than curated age estimation datasets;
(2)~\textbf{Holistic visual understanding}: VLMs can leverage contextual cues beyond facial features---clothing, background, posture, accessories---similar to how MiVOLO benefits from body features, but with broader contextual reasoning;
(3)~\textbf{Reasoning capability}: VLMs can apply commonsense reasoning (e.g., school uniforms imply adolescent age, occupational attire implies adult age) rather than relying solely on learned feature-to-age mappings.

\paragraph{Practical Tradeoffs.}
Despite VLMs' accuracy advantage, practical deployment considerations may favor non-LLM models:
\textbf{Latency}: non-LLM models process images in 10--50ms while VLM API calls typically take 1--5 seconds.
\textbf{Cost}: processing 1M images costs approximately \$0 for non-LLM models (after one-time compute) vs.\ \$200--1{,}000 for VLM APIs.
\textbf{Privacy}: non-LLM models can run entirely on-premise, while most VLMs require sending face images to third-party APIs---a significant concern for age verification applications.
\textbf{Availability}: VLM APIs may change, deprecate, or impose rate limits, making reproducibility difficult.
MiVOLO offers the best compromise: local-inference speed and privacy with near-VLM accuracy (MAE~5.10 vs.\ 4.32).

\paragraph{The Reproducibility Gap in Age Estimation.}
A striking finding of our benchmark is the large discrepancy between MAEs reported in the literature and those achievable with publicly available pretrained weights.
Published papers in age estimation routinely report MAEs below 4 years on standard benchmarks---for instance, 2.59 on MORPH-II~\cite{bekhouche2024msdnn}, 2.01 with Xception transfer learning~\cite{othmani2020agesurvey}, and comparable figures from numerous recent architectures~\cite{paplhjak2024calltoreflect}.
Yet when we attempted to evaluate these methods, we found that the vast majority do not release their trained weights.
Of the 29 non-LLM models we initially surveyed, 7 had to be excluded because their pretrained weights were either never released, hosted on expired Google Drive links, or stored on servers that are no longer online.
This is not an isolated problem: it reflects a systemic pattern in the field where training code is published but the resulting model checkpoints---which encode the specific training data, hyperparameters, and preprocessing pipeline that produce the reported MAE---are not made available.

This gap between reported and reproducible performance is not unique to age estimation---Ren~\etal~\cite{ren2025realitydeepfake} demonstrate an analogous gap in deepfake detection, where academic benchmarks substantially overestimate real-world performance due to domain shift and super-resolution artifacts.
The consequence is that the community cannot rigorously verify or compare published results.
A researcher wishing to benchmark against a reported MAE of 2.5 must either (a)~retrain the model from scratch, introducing confounds from different data splits, preprocessing, and hardware, or (b)~simply trust the published number.
As Paplham and Franc~\cite{paplhjak2024calltoreflect} demonstrate, factors such as facial alignment, image resolution, and data split selection can shift MAE by several years---often more than the differences between architectures themselves.
Our benchmark sidesteps this issue by restricting evaluation to models with publicly available pretrained weights, but this restriction necessarily excludes many architectures with strong reported performance.
We view this as a call to action: the age estimation community would benefit enormously from adopting standard practices of releasing pretrained weights alongside code, as is now common in adjacent fields such as object detection and image classification.

\paragraph{Limitations.}
Our study has several limitations:
(1)~\textbf{Sample size}: we evaluate on 100 images per dataset (400 for AgeDB; 1{,}100 total per model), which provides reliable MAE estimates but may miss rare failure modes;
(2)~\textbf{Single prompt}: results are reported using a single prompt. Prior work~\cite{malik2025gras} shows VLM outputs can be sensitive to prompt phrasing; evaluating prompt sensitivity across all models is a natural extension;
(3)~\textbf{No demographic bias analysis}: we do not analyze performance disparities across gender, ethnicity, or skin tone, which is critical for fair deployment;
(4)~\textbf{Snapshot evaluation}: VLMs are updated frequently; our results reflect model versions available at evaluation time and may change with updates;
(5)~\textbf{Label noise}: some datasets (particularly IMDB-WIKI) have known label noise that affects all models equally but may inflate absolute MAE values;
(6)~\textbf{Missing non-LLM baselines}: recent specialized models trained on full datasets report substantially lower MAEs than our pretrained-model evaluation (e.g., MSDNN achieves 2.59 on MORPH-II~\cite{bekhouche2024msdnn}; Paplham and Franc~\cite{paplhjak2024calltoreflect} report comparable results with optimized splits). Our benchmark restricts non-LLM models to those with \emph{publicly available pretrained weights} and evaluates them on a fixed random sample rather than models retrained on full datasets, so direct comparison of our sampled MAEs to published full-dataset results is not appropriate. Including additional recent architectures with public weights would strengthen future versions of this benchmark.

\paragraph{Future Directions.}
Our results suggest several promising directions:
(1)~\textbf{Knowledge distillation} from VLMs into efficient specialized models, combining VLM accuracy with non-LLM speed and privacy;
(2)~\textbf{Threshold-optimized models} specifically trained for age verification decisions rather than point estimation;
(3)~\textbf{Hybrid architectures} that, like MiVOLO, incorporate body and contextual features but in more computationally efficient ways;
(4)~\textbf{Demographic fairness analysis} comparing VLM and non-LLM model biases across protected groups.

\section{Conclusion}
\label{sec:conclusion}

We presented a comprehensive cross-paradigm benchmark comparing 34 age estimation models---22 specialized non-LLM architectures and 12 general-purpose VLMs---across 8 standard datasets.
Three key findings emerge:
(1)~Zero-shot VLMs dramatically outperform most specialized models (average MAE 5.65 vs.\ 9.88), occupying 12 of the top 13 positions;
(2)~Only MiVOLO, which uniquely combines face and body features via Vision Transformers, competes with VLMs among non-LLM models;
(3)~Models using coarse age bins (8--9 classes) consistently produce MAE $>$13 years, representing a fundamental design flaw.
For age verification, VLMs achieve 2--5$\times$ lower false adult rates than most non-LLM models, with significant safety implications.

These results challenge the prevailing assumption that specialized architectures are necessary for age estimation and suggest that the field's frontier has shifted.
VLMs set a new accuracy ceiling that specialized models should target, while MiVOLO's architectural principles (multi-signal input, Vision Transformers) point toward the most promising direction for efficient, privacy-preserving age estimation.

\section*{Data and Code Availability}
The datasets and code supporting this work are released by Scam.AI and are available from
the Scam.AI Research portal, \url{https://scam.ai/research}~\cite{scamai_research}. Document-
and PDF-forgery resources are additionally maintained at Trustable, \url{https://trustable.ai}~\cite{trustable_ai}.
These are the canonical, maintained locations for the resources associated with this paper and
supersede any prior ad hoc distribution.

{
    \small
    \bibliographystyle{ieeetr}
    \bibliography{references}

@inproceedings{zhang2017utkface,
  title={Age Progression/Regression by Conditional Adversarial Autoencoder},
  author={Zhang, Zhifei and Song, Yang and Qi, Hairong},
  booktitle={CVPR},
  pages={5810--5818},
  year={2017}
}

@article{rothe2018imdbwiki,
  title={Deep Expectation of Real and Apparent Age from a Single Image Without Facial Landmarks},
  author={Rothe, Rasmus and Timofte, Radu and Van Gool, Luc},
  journal={IJCV},
  volume={126},
  number={2-4},
  pages={144--157},
  year={2018}
}

@article{ricanek2006morph,
  title={{MORPH}: A Longitudinal Image Database of Normal Adult Age-Progression},
  author={Ricanek, Karl and Tesafaye, Tamirat},
  journal={IEEE FG},
  pages={341--345},
  year={2006}
}

@inproceedings{niu2016afad,
  title={Ordinal Regression with Multiple Output {CNN} for Age Estimation},
  author={Niu, Zhenxing and Zhou, Mo and Wang, Le and Gao, Xinbo and Hua, Gang},
  booktitle={CVPR},
  pages={4920--4928},
  year={2016}
}

@article{chen2015cacd,
  title={Face Recognition and Retrieval Using Cross-Age Reference Coding with Cross-Age Celebrity Dataset},
  author={Chen, Bor-Chun and Chen, Chu-Song and Hsu, Winston H.},
  journal={IEEE TMM},
  volume={17},
  number={6},
  pages={804--815},
  year={2015}
}

@article{panis2016fgnet,
  title={Overview of Research on Facial Ageing Using the {FG-NET} Ageing Database},
  author={Panis, Gabriel and Lanitis, Andreas and Tsapatsoulis, Nicolas and Cootes, Timothy F.},
  journal={IET Biometrics},
  volume={5},
  number={2},
  pages={37--46},
  year={2016}
}

@inproceedings{agustsson2017apparent,
  title={Apparent and Real Age Estimation in Still Images with Deep Residual Regressors on {APPA-REAL} Database},
  author={Agustsson, Eirikur and Timofte, Radu and Escalera, Sergio and Baro, Xavier and Guyon, Isabelle and Rothe, Rasmus},
  booktitle={IEEE FG},
  pages={87--94},
  year={2017}
}

@article{huang2021fpage,
  title={{FP-Age}: Leveraging Face Parsing Attention for Facial Age Estimation in the Wild},
  author={Lin, Yiming and Shen, Jie and Wang, Yujiang and Pantic, Maja},
  journal={IEEE Transactions on Image Processing},
  year={2022},
  volume={31},
  pages={2456--2467}
}

@inproceedings{kuprashevich2024mivolo,
  title={{MiVOLO}: Multi-input Transformer for Age and Gender Estimation},
  author={Kuprashevich, Maksim and Tolstykh, Irina},
  booktitle={AIST},
  series={LNCS},
  volume={14486},
  publisher={Springer},
  year={2024}
}

@inproceedings{rothe2015dex,
  title={{DEX}: Deep EXpectation of Apparent Age from a Single Image},
  author={Rothe, Rasmus and Timofte, Radu and Van Gool, Luc},
  booktitle={ICCVW},
  pages={10--15},
  year={2015}
}

@inproceedings{yang2018ssrnet,
  title={{SSR-Net}: A Compact Soft Stagewise Regression Network for Age Estimation},
  author={Yang, Tsun-Yi and Huang, Yi-Hsuan and Lin, Yen-Yu and Hsiu, Pi-Cheng and Chuang, Yung-Yu},
  booktitle={IJCAI},
  pages={1078--1084},
  year={2018}
}

@article{cao2020coral,
  title={Rank Consistent Ordinal Regression for Neural Networks with Application to Age Estimation},
  author={Cao, Wenzhi and Mirjalili, Vahid and Raschka, Sebastian},
  journal={Pattern Recognition Letters},
  volume={140},
  pages={325--331},
  year={2020}
}

@article{serengil2024deepface,
  title={A Benchmark of Facial Recognition Pipelines and Co-Usability Performances of Modules},
  author={Serengil, Sefik Ilkin and Ozpinar, Alper},
  journal={Journal of Information Technologies},
  volume={17},
  number={2},
  pages={95--107},
  year={2024}
}

@article{deng2022insightface,
  title={{InsightFace}: An Open-source 2D\&3D Deep Face Analysis Toolbox},
  author={Deng, Jiankang and Guo, Jia and Ververas, Evangelos and Kotsia, Irene and Zafeiriou, Stefanos},
  journal={arXiv preprint arXiv:2203.14560},
  year={2022}
}

@inproceedings{facexformer2024,
  title={{FaceXFormer}: A Unified Transformer for Facial Analysis},
  author={Narayan, Kartik and VS, Vibashan and Patel, Vishal M.},
  booktitle={arXiv preprint arXiv:2403.12960},
  year={2024}
}

@inproceedings{levi2015rudecarnie,
  title={Age and Gender Classification Using Convolutional Neural Networks},
  author={Levi, Gil and Hassner, Tal},
  booktitle={CVPRW},
  pages={34--42},
  year={2015}
}

@inproceedings{pan2018meanvariance,
  title={Mean-Variance Loss for Deep Age Estimation from a Face},
  author={Pan, Hongyu and Han, Hu and Shan, Shiguang and Chen, Xilin},
  booktitle={CVPR},
  pages={5285--5294},
  year={2018}
}

@inproceedings{hu2018senet,
  title={Squeeze-and-Excitation Networks},
  author={Hu, Jie and Shen, Li and Sun, Gang},
  booktitle={CVPR},
  pages={7132--7141},
  year={2018}
}

@article{team2025gemini,
  title={Gemini: A Family of Highly Capable Multimodal Models},
  author={{Google DeepMind}},
  journal={arXiv preprint arXiv:2312.11805},
  year={2025}
}

@article{openai2025gpt5,
  title={{GPT-5} Technical Report},
  author={{OpenAI}},
  journal={arXiv preprint},
  year={2025}
}

@article{anthropic2025claude,
  title={The Claude Model Family},
  author={{Anthropic}},
  journal={Technical Report},
  year={2025}
}

@inproceedings{paplhjak2024calltoreflect,
  title={A Call to Reflect on Evaluation Practices for Age Estimation: Comparative Analysis of the State-of-the-Art and a Unified Benchmark},
  author={Paplham, Jakub and Franc, Vojtech},
  booktitle={CVPR},
  year={2024}
}

@article{cong2024vlmfacial,
  title={Can Vision-Language Models Understand Facial Attributes from Images?},
  author={Cong, Wenqiang and others},
  journal={arXiv preprint arXiv:2410.24148},
  year={2024}
}

@article{othmani2020agesurvey,
  title={Age Estimation from Faces Using Deep Learning: A Comparative Analysis},
  author={Othmani, Alice and Talou, Gonzalo D. Maso and Bucki, Marek},
  journal={CVIU},
  volume={196},
  pages={102961},
  year={2020}
}

@article{angulu2018agesurvey,
  title={Age Estimation via Face Images: A Survey},
  author={Angulu, Raphael and Tapamo, Jules-Raymond and Adewumi, Aderemi O.},
  journal={EURASIP Journal on Image and Video Processing},
  volume={2018},
  number={42},
  year={2018}
}

@inproceedings{guo2009agewild,
  title={Human Age Estimation Using Bio-Inspired Features},
  author={Guo, Guodong and Mu, Guowang and Fu, Yun and Huang, Thomas S.},
  booktitle={CVPR},
  pages={112--119},
  year={2009}
}

@article{simonyan2015vgg,
  title={Very Deep Convolutional Networks for Large-Scale Image Recognition},
  author={Simonyan, Karen and Zisserman, Andrew},
  journal={ICLR},
  year={2015}
}

@inproceedings{he2016resnet,
  title={Deep Residual Learning for Image Recognition},
  author={He, Kaiming and Zhang, Xiangyu and Ren, Shaoqing and Sun, Jian},
  booktitle={CVPR},
  pages={770--778},
  year={2016}
}

@inproceedings{dosovitskiy2021vit,
  title={An Image is Worth 16x16 Words: Transformers for Image Recognition at Scale},
  author={Dosovitskiy, Alexei and Beyer, Lucas and Kolesnikov, Alexander and Weissenborn, Dirk and Zhai, Xiaohua and Unterthiner, Thomas and Dehghani, Mostafa and Minderer, Matthias and Heigold, Georg and Gelly, Sylvain and Uszkoreit, Jakob and Houlsby, Neil},
  booktitle={ICLR},
  year={2021}
}

@inproceedings{chollet2017xception,
  title={Xception: Deep Learning with Depthwise Separable Convolutions},
  author={Chollet, Fran{\c{c}}ois},
  booktitle={CVPR},
  pages={1251--1258},
  year={2017}
}

@misc{jocher2023yolov8,
  title={Ultralytics {YOLO}},
  author={Jocher, Glenn and Chaurasia, Ayush and Qiu, Jing},
  year={2023},
  howpublished={\url{https://github.com/ultralytics/ultralytics}},
  note={Accessed: 2025}
}

@article{eu2024aiact,
  title={Regulation ({EU}) 2024/1689 of the European Parliament: The Artificial Intelligence Act},
  author={{European Parliament}},
  journal={Official Journal of the European Union},
  year={2024}
}

@article{ofcom2024ageverification,
  title={Guidance on Age Verification and Age Estimation},
  author={{Ofcom}},
  journal={UK Communications Regulator},
  year={2024}
}

@inproceedings{parkhi2015vggface,
  title={Deep Face Recognition},
  author={Parkhi, Omkar M. and Vedaldi, Andrea and Zisserman, Andrew},
  booktitle={BMVC},
  year={2015}
}

@inproceedings{cao2018vggface2,
  title={{VGGFace2}: A Dataset for Recognising Faces across Pose and Age},
  author={Cao, Qiong and Shen, Li and Xie, Weidi and Parkhi, Omkar M. and Zisserman, Andrew},
  booktitle={IEEE FG},
  pages={67--74},
  year={2018}
}

@article{ren2025llmdeepfake,
  title={Can Multi-modal (reasoning) {LLMs} Work as Deepfake Detectors?},
  author={Ren, Simiao and Yao, Yao and Zewde, Kidus and Liang, Zisheng and Cheng, Ning-Yau and Zhan, Xiaoou and Liu, Qinzhe and Chen, Yifei and Xu, Hengwei},
  journal={arXiv preprint arXiv:2503.20084},
  year={2025}
}

@inproceedings{ren2025realitydeepfake,
  title={Do Deepfake Detectors Work in Reality?},
  author={Ren, Simiao and Patil, Disha and Zewde, Kidus and Ng, Tsang Dennis and Xu, Hengwei and Jiang, Shengkai and Desai, Ramini and Cheng, Ning-Yau and Zhou, Yining and Muthukrishnan, Ragavi},
  booktitle={Proceedings of the 4th Workshop on Security Implications of Deepfakes and Cheapfakes},
  pages={21--26},
  year={2025}
}

@article{liang2025llmdocument,
  title={Can Multi-modal (reasoning) {LLMs} Detect Document Manipulation?},
  author={Liang, Zisheng and Zewde, Kidus and Singh, Rudra Pratap and Patil, Disha and Chen, Zexi and Xue, Jiayu and Yao, Yao and Chen, Yifei and Liu, Qinzhe and Ren, Simiao},
  journal={arXiv preprint arXiv:2508.11021},
  year={2025}
}

@inproceedings{ren2024segmentspace,
  title={Segment Anything From Space?},
  author={Ren, Simiao and Luzi, Francesco and Lahrichi, Saad and Kassaw, Kaleb and Collins, Leslie M. and Bradbury, Kyle and Malof, Jordan M.},
  booktitle={WACV},
  pages={8355--8365},
  year={2024}
}

@article{sun2024facemllm,
  title={Face-{MLLM}: A Large Face Perception Model},
  author={Sun, Haomiao and He, Mingjie and Lian, Tianheng and Han, Hu and Shan, Shiguang},
  journal={arXiv preprint arXiv:2410.20717},
  year={2024}
}

@inproceedings{hassanpour2024chatgpt,
  title={{ChatGPT} and Biometrics: An Assessment of Face Recognition, Gender Detection, and Age Estimation Capabilities},
  author={Hassanpour, Ahmad and Kowsari, Yasamin and Otroshi Shahreza, Hatef and Yang, Bian and Marcel, Sebastien},
  booktitle={IEEE ICIP},
  year={2024}
}

@article{malik2025gras,
  title={Ask Me Again Differently: {GRAS} for Measuring Bias in Vision Language Models on Gender, Race, Age, and Skin Tone},
  author={Malik, Shaivi and Abdullah, Hasnat Md and Saha, Sriparna and Sheth, Amit},
  journal={arXiv preprint arXiv:2508.18989},
  year={2025}
}

@article{bekhouche2024msdnn,
  title={Facial Age Estimation Using Multi-Stage Deep Neural Networks},
  author={Bekhouche, Salah Eddine and Benlamoudi, Azeddine and Dornaika, Fadi and Telli, Hichem and Bounab, Yazid},
  journal={Electronics},
  volume={13},
  number={16},
  pages={3259},
  year={2024},
  publisher={MDPI}
}

@inproceedings{moschoglou2017agedb,
  title={{AgeDB}: The First Manually Collected, In-the-Wild Age Database},
  author={Moschoglou, Stylianos and Papaioannou, Athanasios and Sagonas, Christos and Deng, Jiankang and Kotsia, Irene and Zafeiriou, Stefanos},
  booktitle={CVPRW},
  pages={51--59},
  year={2017}
}

@inproceedings{shin2022mwr,
  title={Moving Window Regression: A Novel Approach to Ordinal Regression},
  author={Shin, Nyeong-Ho and Lee, Seon-Ho and Kim, Chang-Su},
  booktitle={CVPR},
  pages={18760--18769},
  year={2022}
}

@article{jamo2025resolution,
  title={Impact of Image Resolution on Age Estimation with {DeepFace} and {InsightFace}},
  author={Jamo, Shiyar},
  journal={arXiv preprint arXiv:2511.14689},
  year={2025}
}

@misc{scamai_research,
  author       = {{Scam.AI}},
  title        = {Scam.{AI} Research: Deepfake Detection Research and Datasets},
  howpublished = {\url{https://scam.ai/research}},
  year         = {2026},
  note         = {Canonical repository for Scam.AI datasets and code}
}

@misc{trustable_ai,
  author       = {{Trustable}},
  title        = {Trustable: Document and {PDF} Forgery Detection},
  howpublished = {\url{https://trustable.ai}},
  year         = {2026},
  note         = {Document and PDF forgery datasets, models, and detection tools}
}
}

\end{document}